%% file: main.tex
\documentclass[10pt,twocolumn,letterpaper]{article}

\usepackage{cvpr}              


\definecolor{cvprblue}{rgb}{0.21,0.49,0.74}
\pdfmapfile{+times.map}
\usepackage{newtxtext} 
\usepackage[pagebackref,breaklinks,colorlinks,linkcolor=red,citecolor=cvprblue]{hyperref}
\usepackage{multirow}
\usepackage{subcaption}

\usepackage{algorithm}

\usepackage{algpseudocode}
\usepackage{amsmath}
\usepackage{makecell}
\usepackage[accsupp]{axessibility}

\def\paperID{985} 
\def\confName{CVPR}
\def\confYear{2025}

\title{Not All Parameters Matter: Masking  Diffusion Models for Enhancing Generation Ability}

\author{Lei Wang\textsuperscript{1},\,  Senmao Li\textsuperscript{1},\, Fei Yang\textsuperscript{1}$^\dagger$,\, Jianye Wang\textsuperscript{1},\, Ziheng Zhang\textsuperscript{1}\\ Yuhan Liu\textsuperscript{1},\, Yaxing Wang\textsuperscript{1,2},\, Jian Yang\textsuperscript{1}$^\dagger$\\
\textsuperscript{1}{PCA Lab, VCIP, College of Computer Science, Nankai University} \quad \textsuperscript{2} {Shenzhen Futian, NKIARI}\\
\texttt{\small \{scitop1998,\ senmaonk,\ feiyangflyhigher\}@gmail.com},\,
\texttt{\small \{yaxing,csjyang\}@nankai.edu.cn}
}

\begin{document}
\maketitle
\renewcommand{\thefootnote}{$\dagger$} 
\footnotetext{Corresponding authors.}
\input{sec/0_abstract}    
\input{sec/1_intro}
\input{sec/2_related_work}

\input{sec/3_method}
\input{sec/4_experiments}
\input{sec/5_conclusion}
{
    \small
    \bibliographystyle{ieeenat_fullname}
    \bibliography{main}
}


\end{document}

%% file: sec/0_abstract.tex
\begin{abstract}
The diffusion models, in early stages focus on constructing basic image structures, while the refined details, including local features and textures, are generated in later stages.  Thus the same network layers are forced to learn both structural and textural information simultaneously,  significantly differing from the traditional deep learning architectures (e.g., ResNet or GANs) which  captures or generates the image semantic information at different layers.  This difference inspires us to explore the time-wise diffusion models.  We initially investigate the key contributions of the U-Net parameters to the denoising process and identify that properly zeroing out certain parameters (including large parameters) contributes to denoising, substantially improving the generation quality on the fly. Capitalizing on this discovery, we propose a simple yet effective method—termed “MaskUNet”— that enhances generation quality with   negligible parameter numbers. 
Our method fully leverages timestep- and sample-dependent effective U-Net parameters. To optimize MaskUNet,  we offer two fine-tuning strategies: a training-based approach and a training-free approach, including tailored networks and optimization functions.  In zero-shot inference on the COCO dataset, MaskUNet achieves the best FID score and further demonstrates its effectiveness in downstream task evaluations. Project page: \url{https://gudaochangsheng.github.io/MaskUnet-Page/} 

\end{abstract}

%% file: sec/1_intro.tex
\section{Introduction}
\label{sec:intro}

\begin{figure*}[t]
  \centering
  
  \begin{subfigure}[t]{0.96\linewidth}
    \centering
    \includegraphics[width=\linewidth]{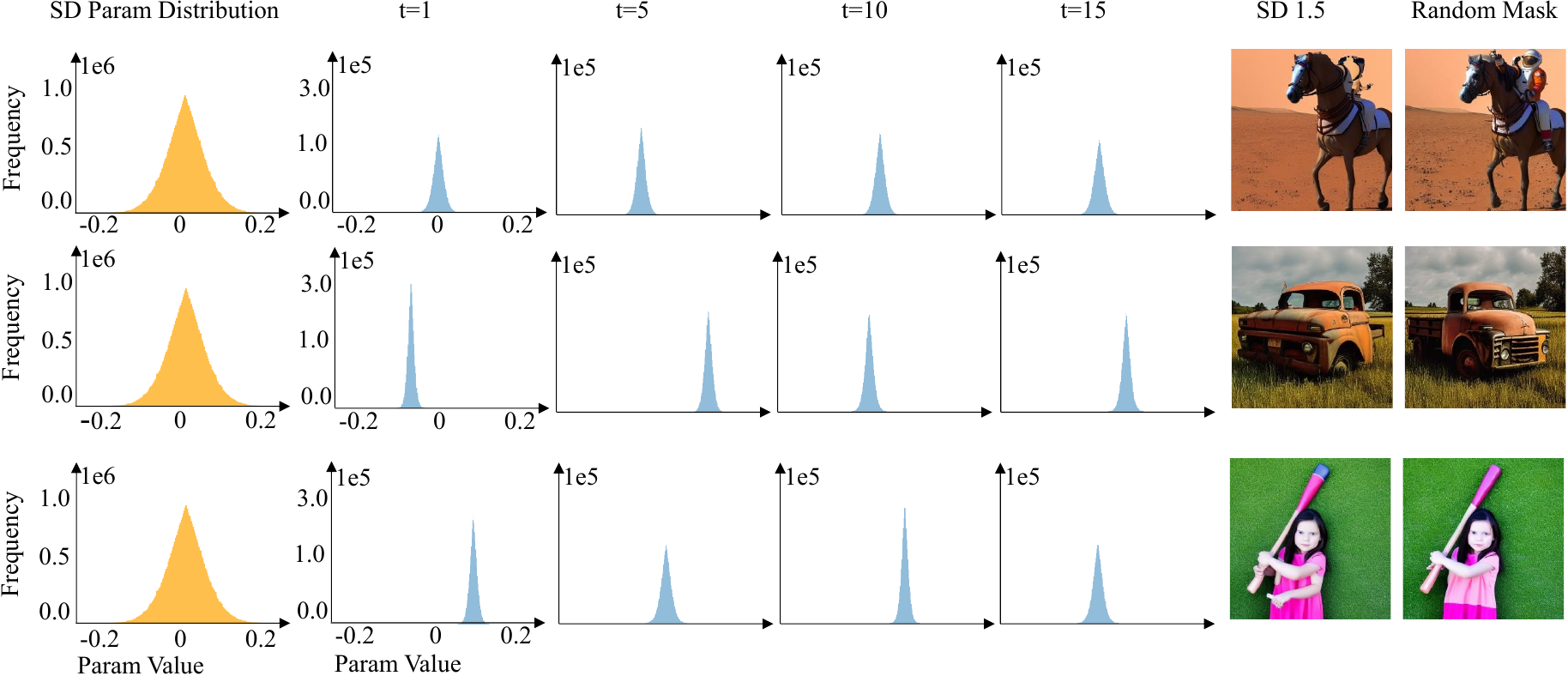}
    \caption{Analysis of parameter distributions and denoising effects across different time steps for Stable Diffusion (SD) 1.5 with and without random masking. The first column shows the parameter distribution of SD 1.5, while the second to fifth columns display the distributions of parameters removed by the random mask. The last two columns compare the generated samples from SD 1.5 and the random mask.}
    \label{fig:onecol}
  \end{subfigure}
  
  \vspace{1em}
  
  \begin{subfigure}[t]{0.57\linewidth}
    \centering
    \includegraphics[width=\linewidth]{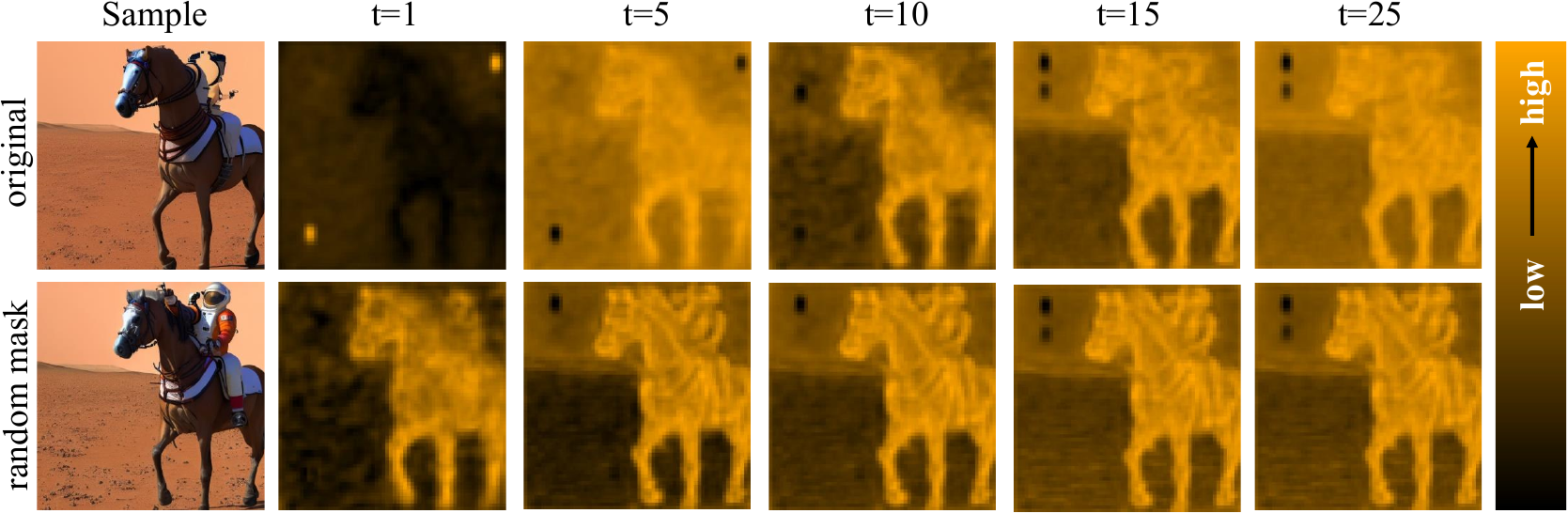}
    \caption{ Comparison of original and random mask results in the denoising process.}
    \label{fig:timestep_vis}
  \end{subfigure}
  \hfill
  \begin{subfigure}[t]{0.40\linewidth}
    \centering
    \includegraphics[width=0.9\linewidth]{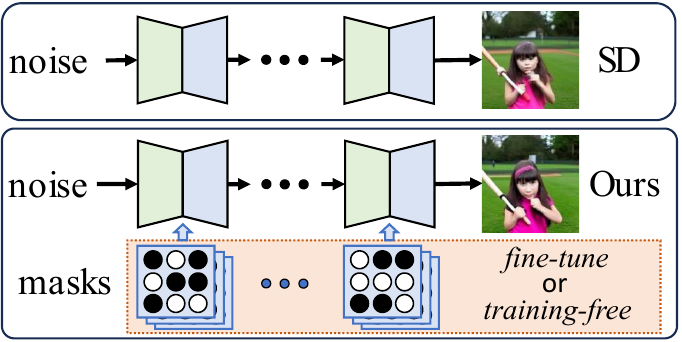}
    \caption{Illustration of our method.}
    \label{fig:abstract_pipeline}
  \end{subfigure}
\vspace{-0.2cm}
  \caption{The motivation of our method.}
  \label{fig:motivation}
\vspace{-0.5cm}
\end{figure*}

Diffusion models~\cite{sohl2015deep,ho2020denoising}, a class of generative models based on iterative denoising processes, have recently gained significant attention as powerful tools for generating high-quality images, videos, and 3D data representations. Text-to-image models, such as stable diffusion (SD)~\cite{rombach2022high}, have successfully applied pre-trained U-Net models to downstream tasks, including personalized text-to-image generation~\cite{ruiz2023dreambooth,li2024photomaker}, relation inversion~\cite{huang2023reversion}, semantic binding~\cite{huang2023t2i,ghosh2024geneval,chefer2023attend,rassin2024linguistic}, and controllable generation~\cite{zhang2023adding,mou2024t2i,zhou2024storydiffusion,du2024learning}. 
The diffusion models, in the early denoising stage,  establish spatial information representing semantic structure, and then widen to the regional details of the elements in the later stage~\cite{choi2022perception,go2023towards}. Therefore, at different inference steps, the diffusion models use the same network paramaters (e.g., a U-Net in SD) to forcibly learn different information: \textit{the global structure and characteristics, and edges and textures etc.}.

However,  the traditional classification models~\cite{szegedy2015going,simonyan2014very,he2016deep,huang2017densely}, such as ResNet~\cite{he2016deep}, they capture the image information (e.g., the structure and semantic features)   at different layers. Typically, the shallow layers  focus on extracting the structure information, while the deeper layers capture higher-level semantic information~\cite{zhou2016learning, selvaraju2017grad,kornblith2019similarity}. Similarly, in traditional generative models~\cite{karras2018progressive,karras2021alias}, the first few layers of the generator control the synthesis of structural information, while the deeper layers represent texture and edge details. Both  classical classification and generative tasks leverage distinct model parts to represent the internal properties of sample, reducing the difficulty of network optimization and enhancing its representational capacity. Distinct from above two classes, the diffusion models use the same parameters to  forcibly learn different information when generating a sample. However, to our best knowledge this  difference of the diffusion U-Net remains largely underexplored.

Beyond the application of diffusion models, in this paper, we are interested in investigating the effectiveness of the pretrained U-Net parameters for the denoising process. To better understand the denoising process, we first present a empirical analysis using a random mask at inference time to examine the generation process of diffusion models,  an area that has received limited prior investigation. As illustrated in Figure~\ref{fig:motivation}~\subref{fig:abstract_pipeline}, we multiply the pre-trained U-Net weights by a random binary UNet-like mask at inference time, ensuring that we have different networks at every time step. This aims to keep the consistency with the traditional network design that  the vary semantic features are modeled at different layers.
As shown in Figure~\ref{fig:motivation}~\subref{fig:onecol} (the second and  last columns),  using certain random masks  enhances  the denoising capability of the U-Net architecture, thereby contributing to a superior output in terms of both fidelity and detail preservation. Further,  we also visualize the corresponding features at different timesteps (see Figure~\ref{fig:motivation}~\subref{fig:timestep_vis}). Compared to the original SD features, the masked backbone features obtain more details and structure information,  improving   the denoising capability. This results  indicate that the generated samples benefit from distinct U-Net weight configurations. 

Based on the above findings, we are interested to  select desire parameters of the diffusion models  which hold the potential to improve sample quality. To achieve this goal, we need to learn a desire binary mask, which zeros out the useless parameters, and retains the desire ones. Naively using a random mask fails to guarantee a good generation result, since the desire mask is related to the denoised sample.  As illustrated in Figure~\ref{fig:motivation} (from third to sixth columns), a vertical examination of each sample reveals that the desire weights differs across samples, indicating that we need a tailored mask to synthesize a high-quality sample.  This insight motivates us to introduce sample dependency in mask generation, allowing the model to better adapt to each prompt's specific needs.

 In this paper, we propel forward with
the introduction of a novel strategy, called \textit{MaskUNet}, which improves the inherent capability of  text-to-image generation  without updating any parameters of the pre-trained U-Net. Specifically, as shown in Figure~\ref{fig:motivation}~\subref{fig:abstract_pipeline}, we introduce a strategy that uses a learnable binary mask to sample parameters from the pre-trained U-Net, thereby obtaining a timestep-dependent and sample-dependent U-Net that emphasizes the importance of parameters sensitive to generation. To efficiently learn the mask, we design two fine-tuning strategies: a training-based approach and a training-free approach. In the training-based approach, a parameter sampler produces timestep-dependent and sample-dependent masks, supervised by diffusion loss. The  parameter sampler is implemented with an MLP, whose parameter count is negligible compared to the pre-trained U-Net. 
 The training-free approach, on the other hand, generates masks directly under the supervision of a reward model~\cite{xu2024imagereward,wu2023human}, eliminating the need for a mask generator compared to the training-based approach.

Compared with existing fine-tuning methods, MaskUNet  aims to tap into the inherent potential of the model, achieving improvements in zero-shot inference accuracy on the COCO 2014~\cite{lin2014microsoft} and COCO 2017~\cite{lin2014microsoft} datasets. We further applied MaskUNet  to downstream tasks, including image customization~\cite{ruiz2023dreambooth,galimage}, video generation~\cite{khachatryan2023text2video}, relation inversion~\cite{huang2023reversion}, and semantic binding~\cite{chefer2023attend,rassin2024linguistic}, to verify its effectiveness. The main contributions of this paper can be summarized as follows:
\begin{itemize}[leftmargin=*]
    \item We conduct an in-depth study of the relationship between parameters in the pre-trained U-Net, samples, and timesteps, revealing the effectiveness of parameter independence, which provides a new perspective for efficient utilization of U-Net parameters.
     \item We propose a novel fine-tuning framework for text-to-image pre-trained diffusion models, called MaskUNet. In this framework, the training-based method optimizes masks through diffusion loss, while the training-free method uses a reward model to optimize masks. The learnable masks enhance U-Net's capabilities while preserving model generalization.
     \item We evaluate MaskUNet  on the COCO dataset and various downstream tasks. Experimental results demonstrate significant improvements in sample quality and substantial performance gains in key metrics.
\end{itemize}

%% file: sec/2_related_work.tex
\section{Related Work}
\label{sec:related-work}
\subsection{Diffusion Models}
Diffusion models \cite{sohl2015deep,ho2020denoising,song2019generative,song2020score} have achieved remarkable success in the field of image generation, but direct computation in pixel space is inefficient. To address this, Latent Diffusion Model (LDM) \cite{rombach2022high} introduces Variational Autoencoders (VAE) to compress images into latent space. Additionally, to tackle iterative denoising during inference, some works have proposed samplers that require fewer steps \cite{songdenoising,zhanggddim,lu2022dpm}, while others have utilized knowledge distillation to reduce sampling steps \cite{chen2024pixart,luo2023latent,nguyen2024swiftbrush,dao2025swiftbrush}. Furthermore, some methods employ structured pruning to accelerate inference \cite{li2023faster,ma2024deepcache}. With the emergence of large-scale image-text datasets \cite{schuhmann2021laion,schuhmann2022laion} and visual language models \cite{radford2021learning,jia2021scaling}, text-to-image generation networks represented by stable diffusion (SD) have found widespread applications, supporting various tasks such as controllable image generation \cite{zhang2023adding,mou2024t2i}, controllable video generation \cite{zhou2024storydiffusion,du2024learning}, and image customization \cite{ruiz2023dreambooth,li2024photomaker,kumari2023multi}.
\subsection{Training-based Models}
Training-based models enhance the U-Net by updating model parameters, typically using the following strategies: introducing trainable modules at specific layers to adapt pretrained weights to new tasks~\cite{ran2024x,mou2024t2i,ye2023ip,guo2024i2v}, selectively fine-tuning a subset of existing parameters~\cite{hu2024sara,guo2020parameter}, or directly updating all parameters. However, these approaches carry a risk of overfitting. Recently, methods like LoRA~\cite{hulora} and DoRA~\cite{liu2024dora} have been proposed, which inject low-rank matrices into pretrained weights to increase model flexibility and mitigate overfitting. However, these methods still adjust the original parameter space, potentially affecting the generalization of the pretrained model. In contrast, our proposed MaskUNet preserves the generalization capacity of the pretrained model by avoiding any updates to the U-Net parameters.
\subsection{Training-free Models}
Training-free models designed to enhance the generative capability of U-Net can be broadly categorized into three main approaches. The first approach focuses on adjusting feature scales~\cite{he2024freestyle,si2024freeu,ma2024surprising}. For instance, FreeU~\cite{si2024freeu} introduces sample-dependent scaling factors for U-Net features and suppresses skip connection features to redistribute feature weights, thereby improving generation quality. The second approach emphasizes optimizing latent codes by leveraging various supervisory methods, such as attention maps~\cite{chefer2023attend,rassin2024linguistic,zhang2024real,agarwal2023star}, noise inversion~\cite{qi2024not}, or reward models~\cite{eyring2024reno}, to strengthen U-Net's generative performance. The third approach centers on optimizing text embeddings~\cite{fengtraining,tunanyan2023multi,Chen_2024_TEBOpt}. For example, Chen \etal~\cite{Chen_2024_TEBOpt} employ balanced text embedding loss to eliminate potential issues within key token embeddings, thus improving generation quality. Unlike these methods, MaskUNet uses a reward model for mask supervision to dynamically select effective U-Net parameters, enhancing its performance.

%% file: sec/3_method.tex
\section{Proposed Method}
\label{sec:method}


\begin{figure*}[t]
    \centering
    \includegraphics[width=0.8\linewidth]{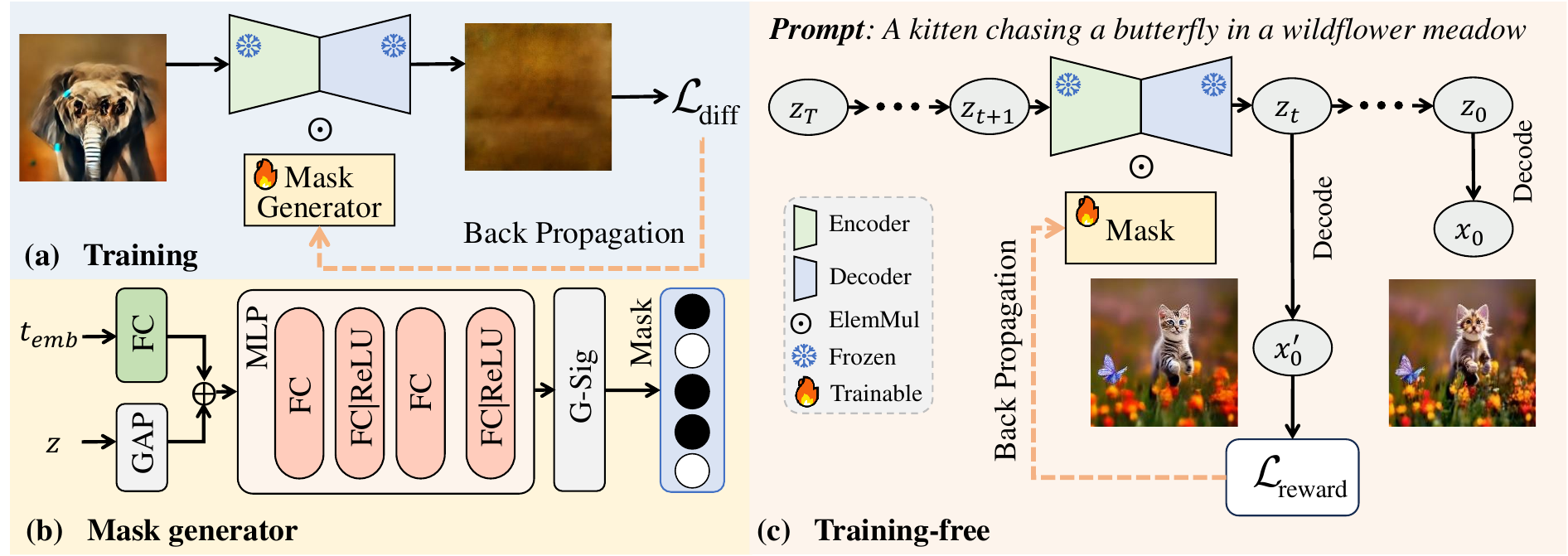}
    \vspace{-0.3cm}
    \caption{The pipeline of the MaskUnet. G-Sig represents the Gumbel-Sigmoid activate function. GAP is global average pooling.}
    \label{fig:method}
\vspace{-0.5cm}
\end{figure*}

The diffusion models use the same parameters to  forcibly learn different information when synthesizing a sample, limiting its  generation adaptability. In this paper, we aim to  learn a timestep-dependent and sample-dependent  mask generation model, which further select the target parameters from the pretrained U-Net, enhancing the pretrained U-Net in the diffusion model. This section first provides an overview of the diffusion model as our foundation (Sec.~\ref{sec:preliminary}), followed by methods for enhancing U-Net through fine-tuning with training (Sec.~\ref{sec:training-based_approach}) and training-free (Sec.~\ref{sec:training-free_approach}) approaches.
\subsection{Preliminary}
\label{sec:preliminary}
Diffusion models add noise to data, creating a Markov chain that approximates a simple prior (usually Gaussian)~\cite{ho2020denoising}. A neural network is trained to reverse this process, starting from noise and progressively denoising to recover the original data, learning to extract useful information at each step.

In the Latent Diffusion Model (LDM)~\cite{rombach2022high}, the diffusion process occurs in a lower-dimensional latent space instead of pixel space, offering significantly improved computational efficiency. The training objective of LDM can be formulated as minimizing the following loss function:
\begin{equation}
    \mathcal{L}_{\mathrm{LDM}}=\mathbb{E}_{\varepsilon, z, t}\left[\operatorname{MSE}\left(\varepsilon_{\theta}\left(z_t, t\right), \varepsilon\right)\right],
\end{equation}
where $\operatorname{MSE}(\cdot)$ denotes the mean squared error, $\varepsilon \sim \mathcal{N}(0, I)$ is noise from a standard Gaussian distribution, $z_t$ is the noisy latent variable at time step $t$, and $\varepsilon_\theta(z_t, t)$ is the noise predicted by the denoising network, parameterized by $\theta$.

In text-to-image diffusion, an additional prompt $c$ guides image generation for more controllable outputs, so the training objective is:
\begin{equation}
    \mathcal{L}_{\mathrm{diff}}=\mathbb{E}_{\varepsilon, z, t, c}\left[\operatorname{MSE}\left(\varepsilon_{\theta}\left(z_t, t, c\right), \varepsilon\right)\right].
    \label{equ:diff_loss}
\end{equation}

\subsection{Training with Learnable Masks}
\label{sec:training-based_approach}

To exploit the full potential of the model's parameters, we introduce a learnable mask to sample weights from the pre-trained U-Net. We propose a training-based fine-tuning approach to optimize the mask, as shown in Figure~\ref{fig:method}(a). The mask is trained using the diffusion loss defined in Equ.~(\ref{equ:diff_loss}). And a mask is generated by a mask generator, as shown in Figure~\ref{fig:method}(b). Let's define the flattened input feature map $ h \in \mathbb{R}^{B \times N \times C_{in}}$, where $ B $ represents the batch size, $N$ is the number of patches and $ C_{\text{in}} $ represents the number of input channels. The mask generator takes as input both the timestep embedding $t_{\text{emb}} \in \mathbb{R}^{B \times C_1}$ and  the latent codes $z \in \mathbb{R}^{B \times C \times H \times W}$, where $H$ and $W$ represent the height and width of $z$.

We first merge $t_{emb}$ and $z$ as follows: \begin{equation} z^{\prime} = \text{FC}(t_{\text{emb}}) + \text{GAP}(z) \,, \end{equation} where $z^{\prime} \in \mathbb{R}^{B \times C}$ is the merged output, $\text{FC}(\cdot)$ is the fully connected layer, and $\text{GAP}(\cdot)$ is global average pooling. We then apply a 4-layer MLP with 2 ReLU activations to introduce non-linearity: 
\begin{equation} 
\hat{z} = \text{MLP}(z^{\prime}),
\end{equation} 
where $\hat{z} \in \mathbb{R}^{B \times C_2}$ is the MLP output.

To sample the weights, we treat $\hat{z}$ as a binary mask: \begin{equation} m = \sigma\left(\hat{S}; \tau, \delta\right) \,, \end{equation} where $m \in \mathbb{R}^{B \times C_2}$ is the output of the Gumbel-Sigmoid~\cite{geng2020does} activation function $\sigma(\cdot; \tau, \delta)$. The temperature coefficient $\tau \in (0, \infty)$ controls the discreteness of $m$: as $\tau \to 0$, $m$ tends to a binary distribution; as $\tau \to \infty$, $m$ tends to a uniform distribution. The threshold $\delta$ is used to discretize the probability distribution.

Next, we apply the reshaped $m^{\prime} \in \mathbb{R}^{B \times C_{\text{out}} \times C_{\text{in}}}$ to the U-Net's linear layer weight $w \in \mathbb{R}^{C_{\text{out}} \times C_{\text{in}}}$ to obtain the masked weight: 
\begin{equation} \hat{w} = m^{\prime} \odot w \,, 
\end{equation} where $\hat{w} \in \mathbb{R}^{B \times C_{\text{out}} \times C_{\text{in}}}$ is the masked weight, and $\odot$ denotes element-wise multiplication.

Finally, the input $h$ and weight $\hat{w}$ are calculated to obtain the output features,
\begin{equation}
    o=\text{BMM}\left(z, \hat{w}\right) \,,
\end{equation}
where $o \in \mathbb{R}^{B \times N \times C_{out}}$, $\text{BMM}\left(\cdot,\cdot\right)$ represents the batch matrix-matrix multiplication.

By introducing a mask generator,  we expect the pretrained U-Net weights to dynamically adapt to different sample features and timestep embeddings. Notably, this design does not update the U-Net’s parameters; instead, it leverages sample- and timestep-dependent adjustments, allowing the model to selectively activate specific U-Net weights tailored to each input. This approach enhances the flexibility of the pretrained U-Net while preserving the stability of the pretrained structure.

\subsection{Training-Free with Learnable Masks}
\label{sec:training-free_approach}

To further demonstrate the effectiveness of the mask, inspired by ReNO~\cite{eyring2024reno}, we propose a training-free algorithm to guide the optimization of the mask.

As shown in Figure~\ref{fig:method}(c), given the intermediate state $z_t$ of the denoising process, which is obtained by denoising the representation $z_{t+1}$ in the previous step and guided by the prompt $c$, it can be expressed as:
\begin{equation}
    z_t = \varepsilon_{\theta}\left(z_{t+1},t+1,c\right),
    \label{equ:ori_x_t}
\end{equation}
where $\varepsilon_{\theta}(\cdot,\cdot,\cdot)$ is the pretrained U-Net, and $\theta$ represents its parameters. 
Similar to the training-based approach, we introduce the mask $m$ to apply to parameter $\theta$, \textit{i.e.}, $\theta^{\prime} \gets \theta \odot m$. The key difference is that $m$ does not rely on the generator.  Therefore, Equ.~(\ref{equ:ori_x_t}) can be rewritten as:
\begin{equation}
    z_t = \varepsilon_{\theta^{\prime}}\left(z_{t+1},t+1,c\right).
\end{equation}
Next, $z_t$ is decoded into pixel space through the VAE to obtain ${x_0}^{\prime}$. Using ${x_0}^{\prime}$ and the prompt $c$, it is fed into the reward model to calculate the loss. The reward loss is then backpropagated to update the mask parameters, improving the consistency between the generated image and the prompt.  The reward loss can be formulated as:
\begin{equation}
    \mathcal{L}_{\text{reward}}=\sum_{i=1}^n\omega_{i}\Psi_{i}\left({x_0}^{\prime},c\right),
\end{equation}
where $\Psi_{i}(\cdot,\cdot)$ denotes the pre-trained reward model, and $\omega_{i}$ is the balancing factor. In this work, we set $n=2$, $n$ is the number of reward models. We use ImageReward~\cite{xu2024imagereward} and HPSv2~\cite{xu2024imagereward} as the reward models. Please check the full details in Algorithm~\ref{algorithm:1}.
\setlength{\textfloatsep}{0pt}
\begin{algorithm}[t]
\caption{Training-free based Fine-tuning}
\begin{algorithmic}[1]
\State \textbf{Require} prompt $c$, a pretrained unet $\varepsilon_{\theta}$, reward models $\sum_{i=1}^n\Psi_{i}$, balance factor of reward models $\omega_{i}$, optimize the number of iterations $\lambda$, mask logits $l$, temperature factor $\tau$, threshold $\delta$, maximum time step $T$
\State \textbf{Initialize} $m$=1.0, $\tau$=1.0, $\delta$=0.5, $x_{T} \sim \mathcal{N}\left(0,\textbf{I}\right)$
\For{$t=T$ \textbf{to} $0$}
    \For{$k=0$ \textbf{to} $\lambda$}
    \State Get binary mask ${m^{\prime}}_t^k \leftarrow \sigma\left(l_t^k;\tau,\delta\right)$
    \State Apply to pre-training unet $g_{\theta^{\prime}}: \theta^{\prime}\leftarrow \theta \odot {m^{\prime}}_t^k$
    \State Predict noisy latent $z_{t-1} \leftarrow \varepsilon_{\theta^{\prime}}\left(z_t, t, c\right)$
    \State Predict the original latent $z_0 \leftarrow z_{t-1}$
    \State Decode to image space $x_{t}^k \leftarrow z_0$
    \State Reward loss $\mathcal{L}_{\text{reward}} \leftarrow \sum_{i=1}^{n} \omega_{i} \Psi_{i}\left(x_{t}^k, c\right)$
    \State Update mask logits $l_t^{k+1} \leftarrow l_t^k$
    \EndFor
\EndFor
\State \Return $x_{0}^{\lambda}$
\end{algorithmic}
\label{algorithm:1}
\end{algorithm}

%% file: sec/4_experiments.tex
\begin{figure*}[t]
  \centering
   \includegraphics[width=0.92\linewidth]{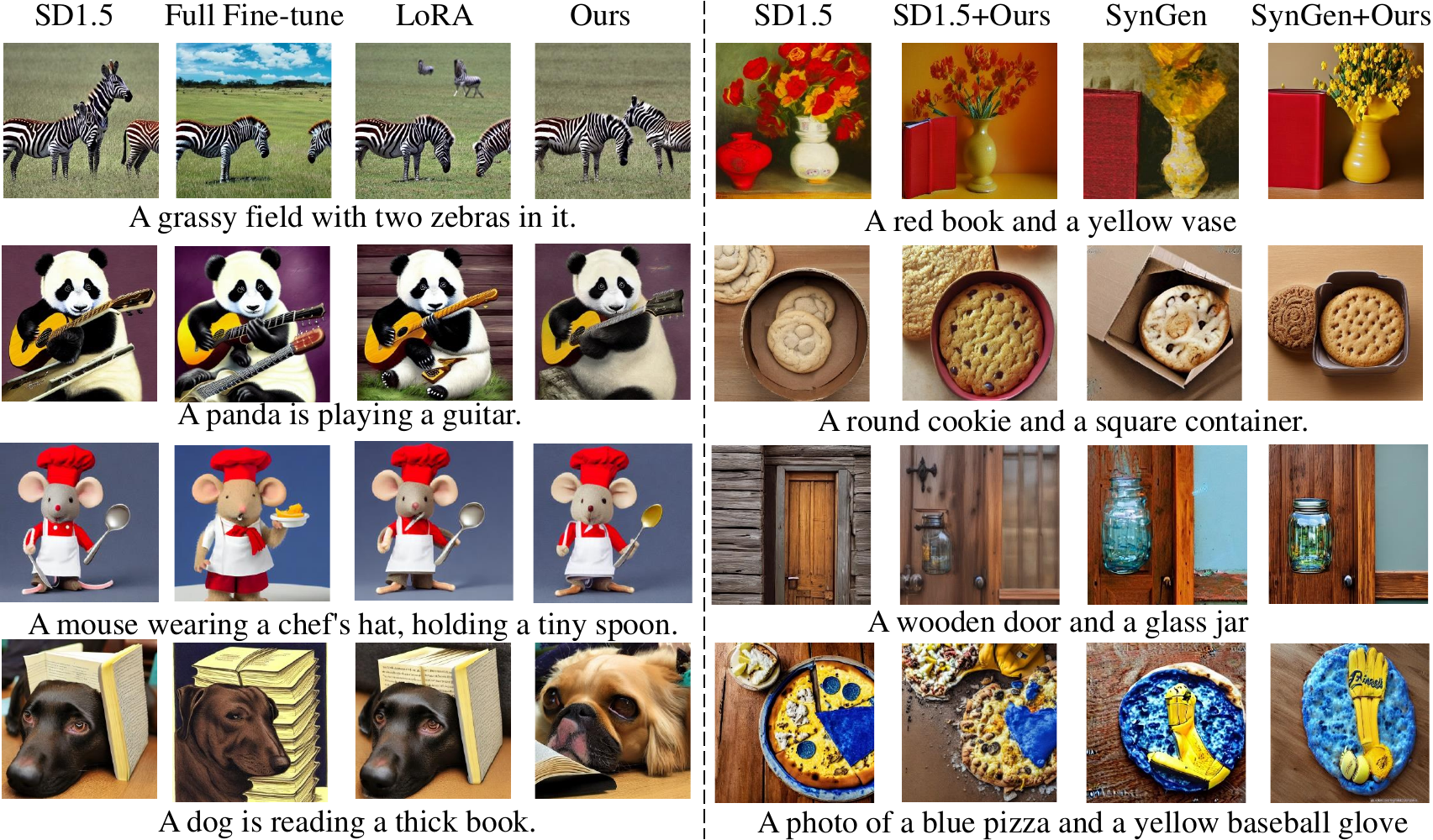}
   \vspace{-0.3cm}
   \caption{Quality results compared to other methods.}
   \label{fig:dataset_vis}
\vspace{-0.5cm}
\end{figure*}

\section{Experiments}
\label{sec:experiment}

\subsection{Experiment Setting}
\noindent\textbf{Datasets and Metrics.} (1) \textit{Training-based approach.} For zero-shot text-to-image generation, we fine-tune the MaskUNet on a subset of the Laion-art (a subset of Laion-5B \cite{schuhmann2022laion}), which contains 20.1k pairs of image and text. To verify the effectiveness of our method, we generated 30k images for COCO 2014 \cite{lin2014microsoft}, 5k images for COCO 2017 \cite{lin2014microsoft} respectively. We evaluate the image quality using Fréchet Inception Distance (FID) \cite{heusel2017gans} and the alignment of the image text using CLIP score \cite{radford2021learning}. (2) \textit{Training-free approach.} We evaluated the effectiveness of MaskUNet on two semantic binding datasets T2I-CompBench \cite{huang2023t2i} and GenEval \cite{ghosh2024geneval}. We use BLIP-VQA score \cite{huang2023t2i} for the evaluation of attribute correspondences and GENEVAL score for the image correctness.

\noindent\textbf{Baselines}. (1) \textit{Training-based approach.} We choose SD 1.5 \cite{rombach2022high}, Full Fine-tune and LoRA \cite{hulora} as baselines to compare with MaskUnet. We also apply MaskUNet to downstream tasks such as image customization, relation inversion, and text-to-video generation. For these tasks,  we use Dreambooth \cite{ruiz2023dreambooth}, Textual Inversion \cite{galimage}, ReVersion \cite{huang2023reversion} and Text2Video-zero \cite{khachatryan2023text2video} as baselines. (2) \textit{Training-free approach.} We select SD 1.5 \cite{rombach2022high}, SD 2.0 \cite{rombach2022high}, SynGen \cite{rassin2024linguistic} and Attend-and-excite \cite{chefer2023attend} as baselines for comparison with MaskUNet.

\noindent\textbf{Implementation Details}.
(1) \textit{Training-based approach.} In our implementation, the learning rate (LR) is set to 1$e$-5, and AdamW \cite{loshchilov2017decoupled} with a weight decay of 1$e$-2 is used as the optimizer.
The training process consists of 12 epochs, with 50 inference steps. 
The classifier-free guidance (CFG) \cite{ho2022classifier} is set to 7.5, and DDIM \cite{zhanggddim} is employed as the sampler. 
(2) \textit{Training-free approach.} The number of iterations $\lambda$ is set to 15. The optimizer used is AdamW \cite{loshchilov2017decoupled}, with an LR of 1$e$-2. We utilize ImageReward \cite{xu2024imagereward} and HPSV2 \cite{wu2023human} as reward models, with equilibrium coefficients set to 1.0 and 5.0, respectively. The number of inference steps is set to 15, with the CFG \cite{ho2022classifier} set to 7.5. The sampler uses DPM-Solver \cite{lu2022dpm}.

\begin{table}[htbp]
  \centering
  \caption{Quantitative results of zero-shot generation on the COCO 2014 and COCO 2017 datasets, with the best results in \textbf{bold}.}
  \vspace{-0.3cm}
  \resizebox{0.999\linewidth}{!}{
    \begin{tabular}{ccccc}
    \toprule
    \multirow{2}[4]{*}{Method} & \multicolumn{2}{c}{COCO 2014} & \multicolumn{2}{c}{COCO 2017} \\
\cmidrule{2-5}          & FID-30k ($\downarrow$)   & CLIP ($\uparrow$)  & FID-5k ($\downarrow$)   & CLIP ($\uparrow$)\\
    \midrule
    SD 1.5 \cite{rombach2022high} & 12.85 & 0.32 & 23.39 & 0.33 \\
    Full Fine-tune  & 14.06 & 0.32 & 24.45 & 0.33 \\
    LoRA \cite{hulora}  & 12.82 & 0.32 & 23.18 & 0.33 \\
    \midrule
    MaskUnet  & \textbf{11.72} & 0.32 & \textbf{21.88} & 0.33 \\
    \bottomrule
    \end{tabular}%
    }
  \label{tab:coco-result}%
\end{table}%

\subsection{Training-based Text-to-image Generation}

\subsubsection{Zero-shot Text-to-image Generation}
Table~\ref{tab:coco-result} presents the zero-shot generation performance of our method and baselines on the COCO 2014 and COCO 2017 datasets. For COCO 2014, MaskUNet improves the FID by 1.13 compared to SD 1.5 \cite{rombach2022high} and by 1.10 over LoRA \cite{hulora}. In contrast, Full Fine-tune shows an increased FID value by 1.21 compared to SD v1.5, indicating a risk of overfitting. A similar trend is observed on the COCO 2017 dataset. In summary, by leveraging the dynamic masking mechanism, MaskUNet effectively enhances the generative performance of the SD \cite{rombach2022high} model.
%

Figure~\ref{fig:dataset_vis} (left) presents the generative results of different methods for various prompts. In the first row, MaskUNet generates realistic and well-aligned images, while other methods either introduce artifacts (e.g., LoRA \cite{hulora}) or display unnecessary background elements due to overfitting (e.g., Full Fine-tune). In the third row, MaskUNet accurately captures the mouse’s attire and pose, a level of consistency that other methods struggle to achieve. Overall, MaskUNet effectively balances image quality and fidelity to prompts, capturing prompt-specific details while maintaining visual coherence across diverse scenes.

\subsubsection{Downstrean tasks}
MaskUnet also has the potential to enhance image quality in a variety of downstream tasks, with evaluations ranging from image customization, relation inversion, and text-to-video generation tasks.

\begin{figure}[t]
  \centering
   \includegraphics[width=0.98\linewidth]{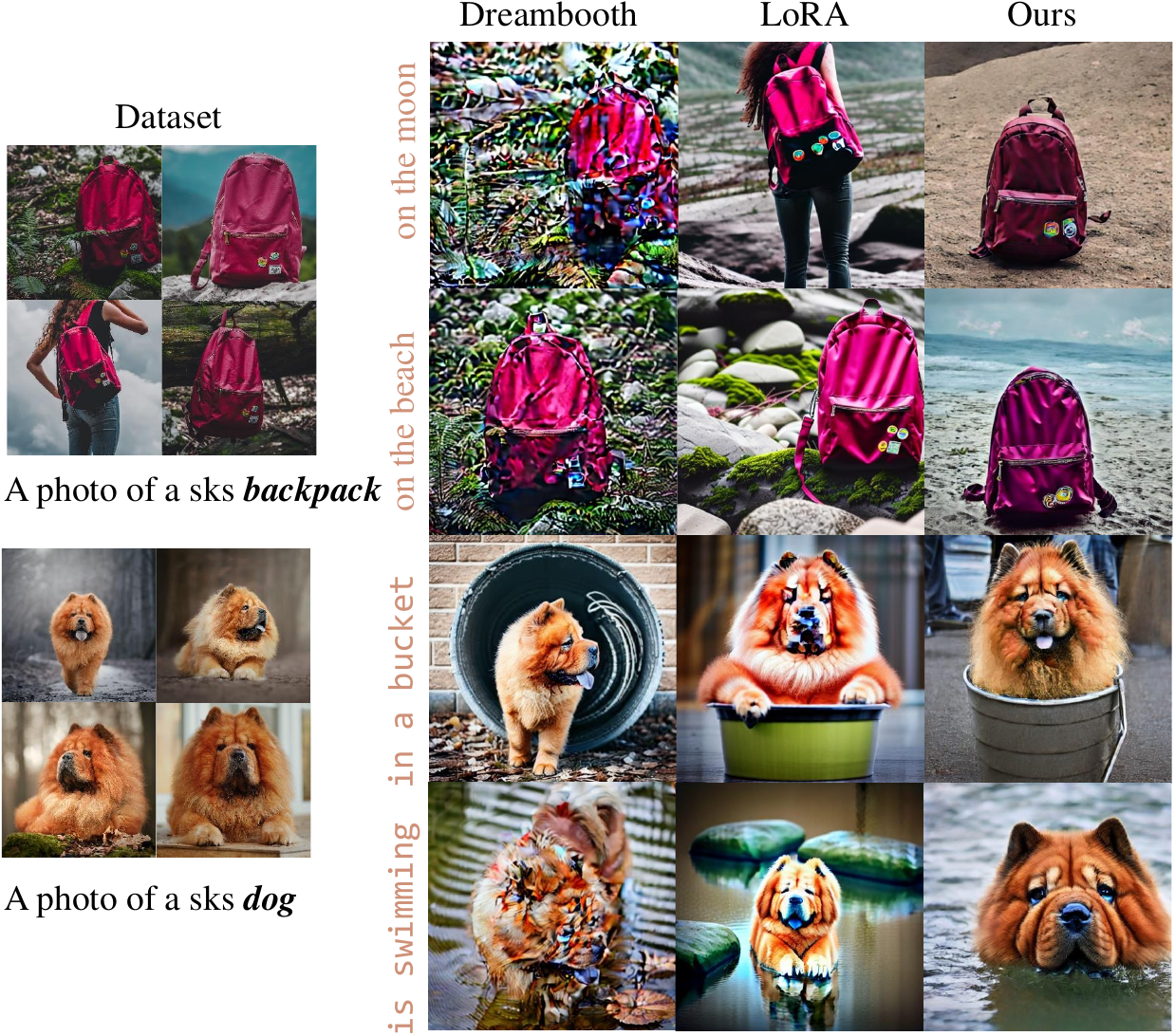}
   \vspace{-0.3cm}
   \caption{Quality results compared to other methods.}
   \label{fig:dreambooth}
\end{figure}

\noindent\textbf{Image Customization.} DreamBooth \cite{ruiz2023dreambooth} is a pioneering method for image customization, which it requires full fine-tuning of the U-Net. We compared the performance of full fine-tuning (DreamBooth), LoRA \cite{hulora}, and MaskUNet. As shown in Figure~\ref{fig:dreambooth}, MaskUNet excels in maintaining subject consistency and background diversity, producing high-quality images across diverse prompts, while DreamBooth and LoRA exhibit overfitting. For example, with a rare prompt combination like “on the moon”, DreamBooth fails to generate coherent images, and LoRA retains unwanted elements from the training set, such as background details. Notably, MaskUNet achieves effective personalization without updating U-Net parameters, demonstrating the untapped potential of the pretrained U-Net.

\begin{figure}[t]
  \centering
   \includegraphics[width=0.98\linewidth]{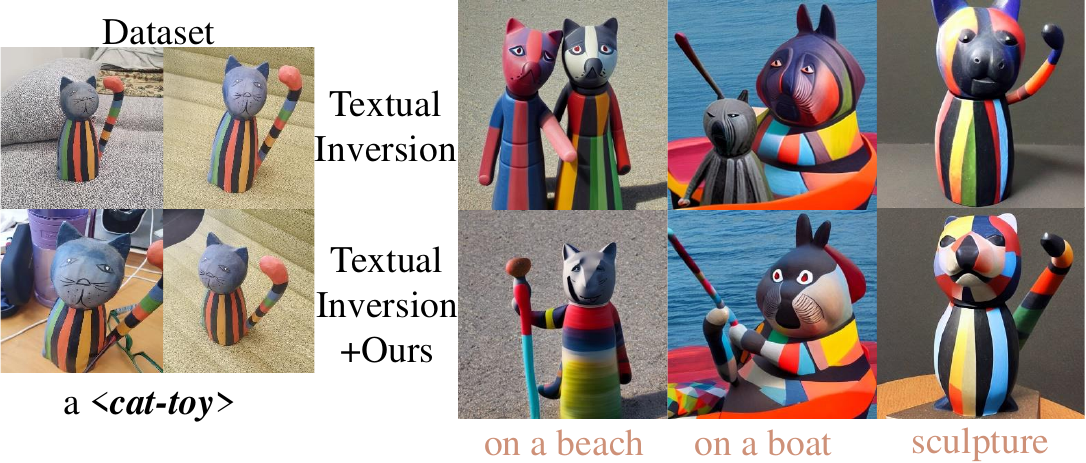}
   \vspace{-0.3cm}
   \caption{Quality results by Textual Inversion~\cite{galimage} with or without mask.}
   \label{fig:textual_inversion}
\end{figure}

Textual Inversion~\cite{galimage} learns text embeddings to capture new concepts and is further enhanced with the introduction of MaskUNet. As shown in Figure~\ref{fig:textual_inversion}, adding mask significantly improves the generation quality of Textual Inversion. For instance, the results in the first and second columns show enhanced sensitivity to quantity, while the third column better preserves subject characteristics, resulting in more accurate outputs.

\begin{figure}[t]
  \centering
   \includegraphics[width=0.98\linewidth]{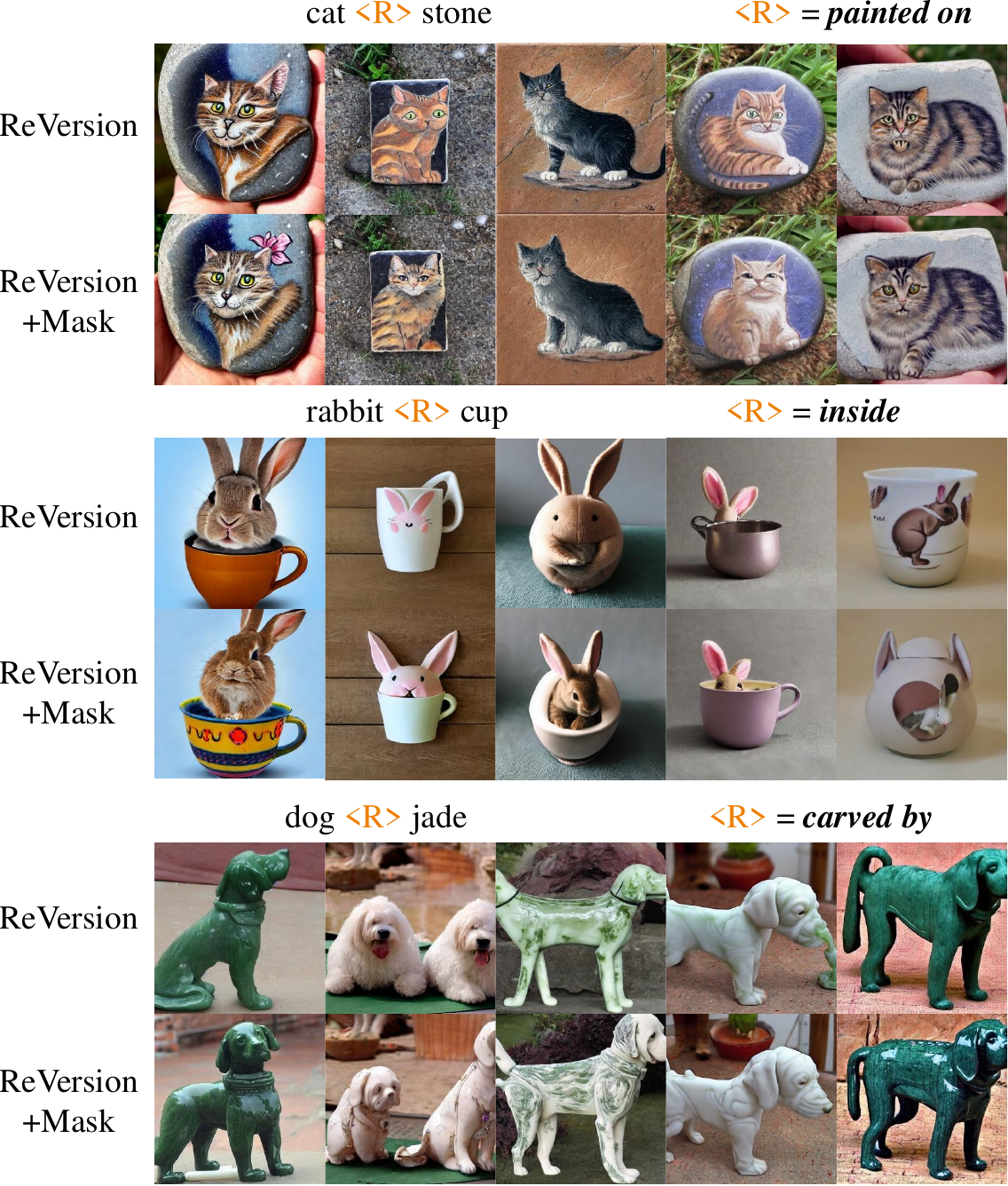}
   \vspace{-0.3cm}
   \caption{Quality results by ReVersion~\cite{huang2023reversion} with or without mask.}
   \label{fig:reversion}
\end{figure}

\noindent\textbf{Relation Inversion.} ReVersion~\cite{huang2023reversion}, a relationship-guided image synthesis method based on SD, can be enhanced by integrating MaskUNet. As shown in Figure~\ref{fig:reversion}, adding mask improves sensitivity to relational embeddings and enhances image fidelity. For instance, with the prompt “inside,” ReVersion might place the rabbit on the surface of the cup or outside it, but with MaskUnet, sensitivity to the “inside” embedding is increased, resulting in images with the correct relational context. Additionally, for prompts like “cat,” adding mask significantly enhances image quality.

\begin{figure*}[t]
  \centering
   \includegraphics[width=0.75\linewidth]{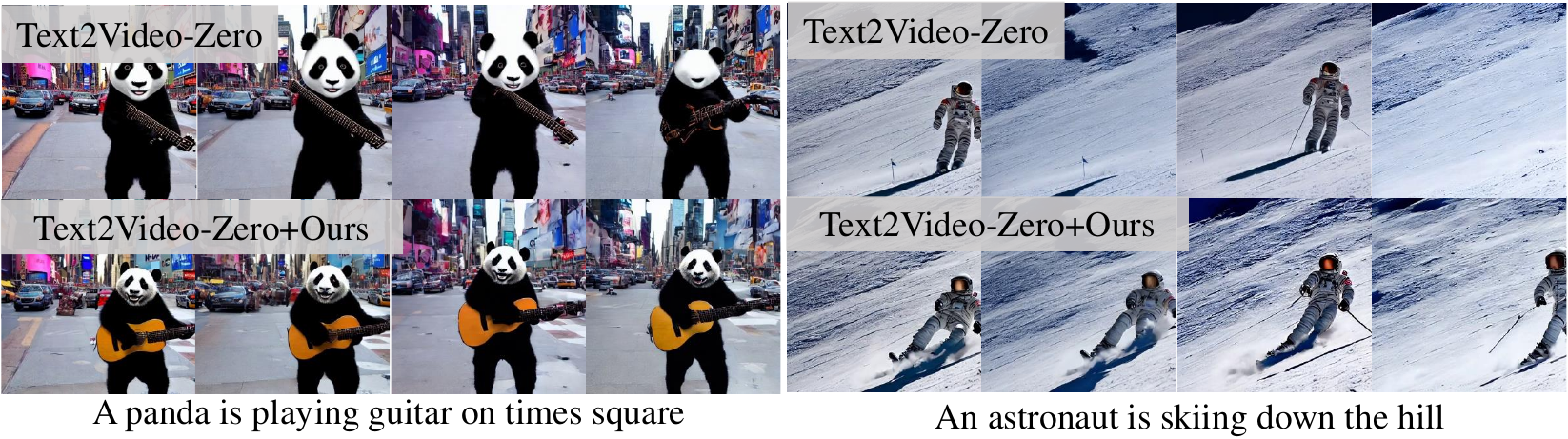}
   \vspace{-0.3cm}
   \caption{Quality results by Text2Video-Zero~\cite{khachatryan2023text2video} with or without mask.}
   \label{fig:text_to_video}
\vspace{-0.4cm}
\end{figure*}

\noindent\textbf{Text-to-video Generation.} Tex2Video-Zero~\cite{khachatryan2023text2video} is a training-free diffusion model for text-to-video generation. By integrating our MaskUnet into Tex2Video-Zero, we can enhance the continuity and consistency of generated videos, as illustrated in Figure~\ref{fig:text_to_video}. For instance, in response to the prompt "A panda is playing guitar on Times Square," the addition of the mask enables the generation of a complete guitar. This indicates that the mask is orthogonal to Tex2Video-Zero, thereby facilitating the production of high-quality content.

\subsection{Training-free Text-to-image Generation}
\noindent\textbf{Semantic Binding.} Table~\ref{tab:t2icompbench} presents the quantitative results of MaskUNet on the T2I-Compbench benchmark. We observe that, compared to SD v1.5, MaskUNet achieves over 7\% improvement across color, shape, and texture categories. When compared to SD 2.0~\cite{rombach2022high}, MaskUNet slightly underperforms in color but surpasses it in the other two categories. To further verify the generalizability of MaskUNet, we applied it to SynGen \cite{rassin2024linguistic}, resulting in over 4\% improvement in all three categories. Similar findings are shown in Table~\ref{tab:geneval} on the GenEval benchmark, where the color attribution score in SynGen increased by 21\% after applying MaskUNet. In summary, our MaskUNet demonstrates robust generalization capabilities in semantic binding tasks.

Figure~\ref{fig:dataset_vis} (right) compares samples generated by different methods to evaluate the effectiveness of MaskUnet in semantic binding tasks. In the first row, adding MaskUnet to SD 1.5 highlights the semantic information of the ``book", while its addition to SynGen enhances the vase's texture from blurry to detailed. In the second row, MaskUnet improves sensitivity to quantity and shape. In the third row, it enhances texture generation. In the last row, MaskUnet enables more accurate adherence to the specified color and object combination. Overall, MaskUnet significantly improves generative quality in semantic binding tasks, demonstrating higher fidelity to prompt specifications.
\begin{table}[htbp]
\vspace{-0.3cm}
  \centering
  \caption{Semantic binding evaluation for T2I-CompBench, with the best results in \textbf{bold}.}
  \vspace{-0.3cm}
  \resizebox{0.94\linewidth}{!}{
    \begin{tabular}{c|c|ccc}
    \toprule
    \multirow{2}[4]{*}{Method} &\multirow{2}[4]{*}{NFE}& \multicolumn{3}{c}{BLIP-VQA} \\
\cmidrule{3-5}     &     & Color ($\uparrow$) & Texture ($\uparrow$) & Shape ($\uparrow$) \\
    \midrule\midrule
    SD 1.5 \cite{rombach2022high} &15& 0.3750  & 0.4159  & 0.3742  \\
    SD 2.0 \cite{rombach2022high} &50& 0.5056  & 0.4922  & 0.4221  \\
    \midrule
    SynGen \cite{rassin2024linguistic} &15& 0.6288  & 0.5796  & 0.3881  \\
    Atten-Exct \cite{chefer2023attend} &50& 0.6400  & 0.5963  & 0.4517  \\
    \midrule
    MaskUNet  &15& 0.4958  & 0.4938  & 0.4529  \\
    SynGen+MaskUNet &15& 0.\textbf{6989}  & \textbf{0.6209}  & \textbf{0.4644}  \\
    \bottomrule
    \end{tabular}%
    }
  \label{tab:t2icompbench}%
  \vspace{-0.4cm}
\end{table}%

\begin{table}[htbp]
\vspace{-0.3cm}
\tabcolsep=0.04cm
  \centering
  \caption{Semantic binding evaluation for GeneVal, with the best results in \textbf{bold}.}
  \vspace{-0.3cm}
  \resizebox{0.999\linewidth}{!}{
    \begin{tabular}{c|cc|cc}
    \toprule
    Model & SD 1.5 \cite{rombach2022high}  & SynGen \cite{rassin2024linguistic} & MaskUNet  & \makecell{SynGen+\\MaskUNet} \\
    \midrule
    Overrall ($\uparrow$) & 0.39    & 0.43  & 0.46  & \textbf{0.50}  \\
    Single ($\uparrow$) & 0.98    & 0.94  & 0.98  & \textbf{0.10}  \\
    Two ($\uparrow$)  & 0.26    & 0.39  & 0.42  & \textbf{0.43}  \\
    Counting ($\uparrow$) & 0.28    & 0.31  & 0.38  & \textbf{0.39}  \\
    Colors ($\uparrow$) & 0.74    & 0.80  & 0.82  & \textbf{0.88}  \\
    Position ($\uparrow$) & 0.02    & 0.06  & 0.06  & \textbf{0.08}  \\
    Color Attri ($\uparrow$) & 0.05   & 0.05  & 0.08  & \textbf{0.26} \\
    \bottomrule
    \end{tabular}%
    }
  \label{tab:geneval}%
  \vspace{-0.5cm}
\end{table}%

\subsection{User Study}
\begin{figure}[t]
  \centering
   \includegraphics[width=1.\linewidth]{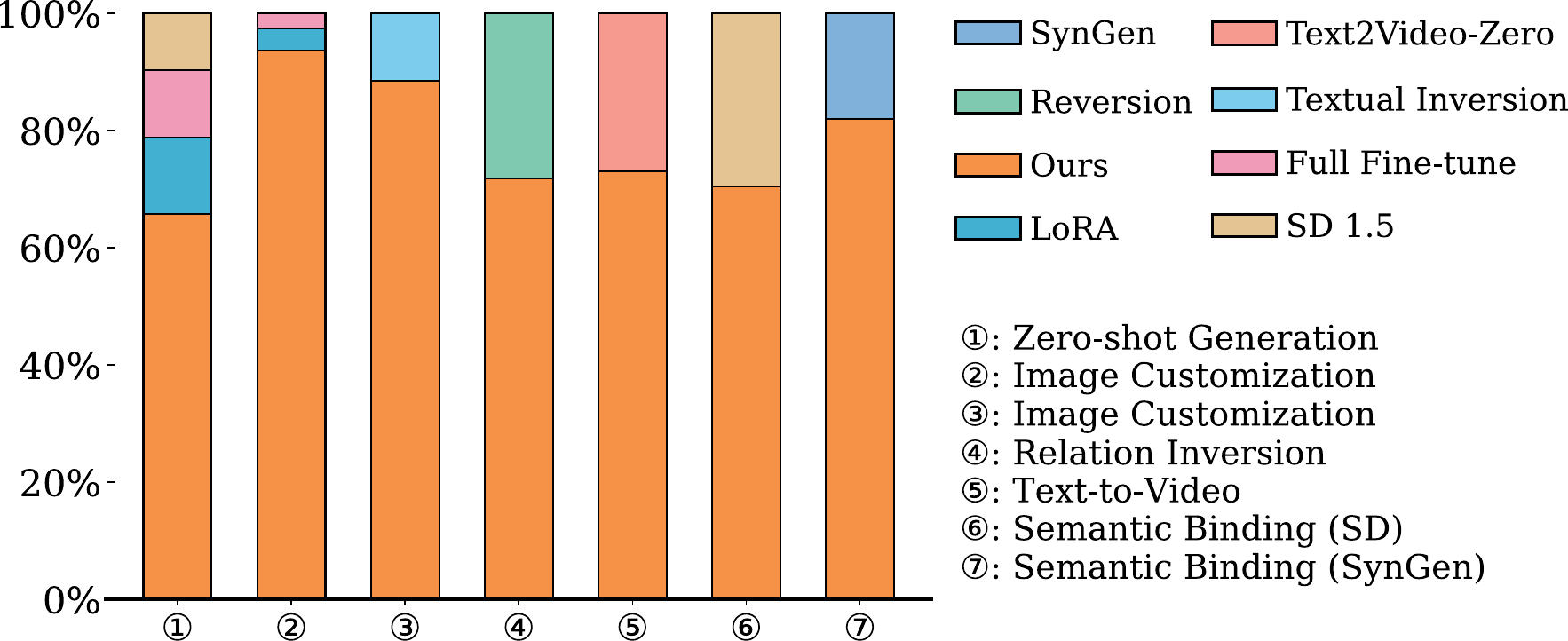}
   \vspace{-0.8cm}
   \caption{Quantitative results compared to other methods.}
   \label{fig:user_study}
\end{figure}
We conducted a study with 26 participants to evaluate image quality and text-image alignment, covering zero-shot generation and downstream tasks. Figure~\ref{fig:user_study} presents the voting results, where the majority of votes favored our method, indicating that our approach effectively enhances the generative capability of SD.
\subsection{Analysis of UNet Weight Masks}
\begin{figure}[tb]
\centering
    \subfloat[Image distribution]{
        \includegraphics[width=0.55\linewidth]{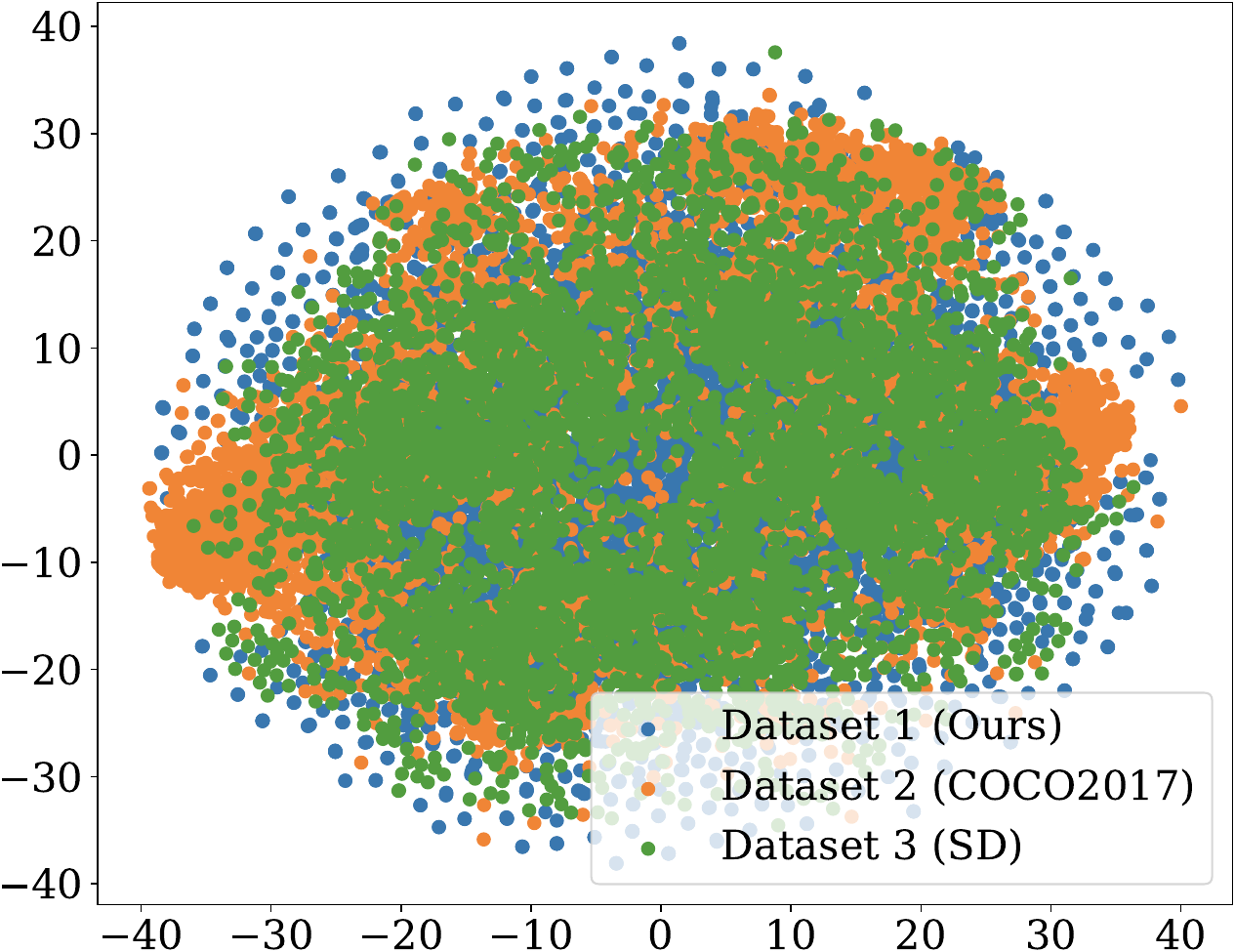}
        \label{fig:sd_mask_coco}
        }
    \subfloat[Mask ratio v.s. FID]{
        \includegraphics[width=0.42\linewidth]{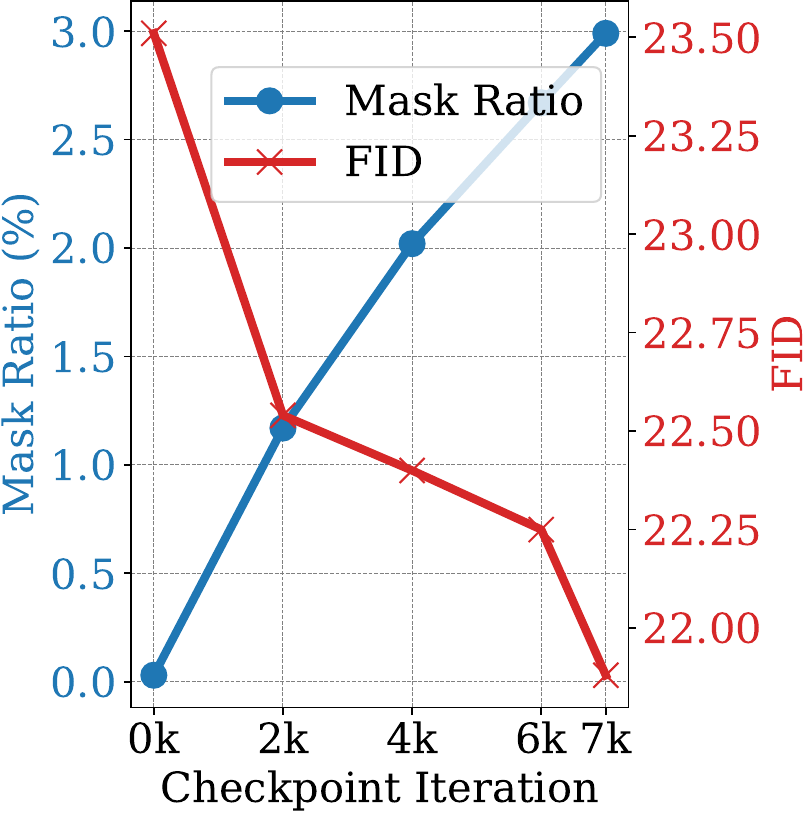}
        \label{fig:ratio_fid}
        }
    \vspace{-0.3cm}
    \caption{(a) Visualization of image distributions for different methods using t-SNE. (b) Relationship between mask ratio and FID across checkpoint iterations.}
    \label{fig:ana-mask}
\end{figure}

As shown in Figure~\ref{fig:ana-mask}~\subref{fig:sd_mask_coco}, we visualized the distributions of images generated by MaskUNet, images generated by SD, and real images from COCO 2017 using the t-SNE \cite{van2008visualizing} dimensionality reduction method. It can be observed that the distribution of images generated by MaskUNet is closer to the real image distribution. Therefore, this reveals the reason why MaskUNet enhances the generalization ability of SD. Then, as shown in Figure~\ref{fig:ana-mask}~\subref{fig:ratio_fid}, we observe that as the number of iterations increases, the mask ratio shows an upward trend, while the FID gradually decreases, indicating that the mask is continuously enhancing the generative capability. It is worth noting that although the overall mask ratio remains constant, the mask locations change dynamically, resulting in a varying distribution of masked parameters (see the \textit{supplementary material}).

\subsection{Ablation Studies}
For the training-based approach, Table~\ref{tab:ablation} shows an ablation study on the effectiveness of different inputs to the mask generator. The MaskUNet, with both timestep embeddings and sample inputs, achieves the lowest FID score. Removing either the timestep embeddings or sample inputs results in higher FID scores, with the SD 1.5 (no mask) performing the worst. All experiments have similar CLIP scores, indicating that the mask primarily improves image quality without significantly affecting semantic alignment.
\begin{table}[htbp]
  \centering
  \caption{Ablation study on the impact of different inputs to the mask generator on COCO 2017.}
  \vspace{-0.2cm}
    \begin{tabular}{ccc}
    \toprule
    Model & FID ($\downarrow$)   & CLIP ($\uparrow$) \\
    \midrule
    MaskUNet  & \textbf{21.88} & \textbf{0.33} \\
    w/o temb & 22.30  & 0.32 \\
    w/o sample & 22.14     & 0.32 \\
    SD 1.5    & 23.39 & 0.33 \\
    \bottomrule
    \end{tabular}%
  \label{tab:ablation}%
  \vspace{-0.4cm}
\end{table}%

%% file: sec/5_conclusion.tex
\section{Conclusion}
\label{sec:conclusion}
This paper proposes MaskUNet, an enhanced method for U-Net parameters in diffusion models. By utilizing learnable binary masks, MaskUNet generates time-step and sample-dependent U-Net parameters during inference. Experimental results demonstrate that MaskUNet significantly enhances the generative capability of U-Net, with improved sample quality observed in the COCO zero-shot task. Additionally, our method outperforms existing approaches in downstream tasks such as image customization, relation inversion, and text-to-video generation. To optimize computational efficiency, we also introduce a mask learning approach that requires no training, and we validate its effectiveness on two semantic binding benchmarks.

\textbf{Limitations.} While dynamic masking enhances model generalization, it does not enable learning of new knowledge. Future work will explore combining this approach with LoRA and extending it to other base models.

\textbf{Acknowledgement.} This work was supported by the National Science Fund of China under Grant Nos, 62361166670 and U24A20330, the ``Science and Technology Yongjiang 20'' key technology breakthrough plan project (2024Z120), the Shenzhen Science and Technology Program (JCYJ20240813114237048), and the Supercomputing Center of Nankai University (NKSC).